\documentclass{article}

\usepackage[numbers]{natbib}
\usepackage[final,main]{neurips_2026}

\usepackage[utf8]{inputenc} %
\usepackage[T1]{fontenc}    %
\usepackage{hyperref}       %
\usepackage{url}            %
\usepackage{booktabs}       %
\usepackage{amsmath}
\usepackage{amsfonts}       %
\usepackage{nicefrac}       %
\usepackage{microtype}      %
\usepackage{xcolor}         %

\usepackage{bm}
\usepackage{graphicx}
\usepackage{wrapfig}
\usepackage{colortbl}
\usepackage{adjustbox}
\usepackage{afterpage}

\title{3D Gaussian Splatting within Self-Learned Neural View-Dependent Color Fields}

\author{%
Yixiao Yang\textsuperscript{1} \quad
Yu-Shen Liu\textsuperscript{1}\thanks{Corresponding author.} \quad
Zhizhong Han\textsuperscript{2} \\
\textsuperscript{1}School of Software, Tsinghua University, Beijing, China \\
\textsuperscript{2}Department of Computer Science, Wayne State University, Detroit, USA \\
\texttt{yangyixi25@mails.tsinghua.edu.cn, liuyushen@tsinghua.edu.cn} \\
\texttt{h312h@wayne.edu}
}

\begin{document}

\maketitle

\begin{abstract}
3D Gaussian Splatting (3DGS) has become a vital tool in novel view synthesis. 3D Gaussian functions with attributes like color, opacity, and scale are stacked together to explicitly model a radiance field. However, discrete 3D Gaussians usually produce poor depth with few geometry details but lots of floaters, which hinders the effective modeling of smooth radiance fields. To address this issue, we propose to learn a radiance field within a continuous view-dependent color field parameterized by neural networks. With continuous constraints in color, the bias on low-frequency signals of neural networks can highly encourage the geometry modeling to contribute to high-frequency color variations on images, getting rid of solely relying on the color attribute itself. Moreover, we impose a constraint to improve the Lipschitz continuity of the neural color field, making large changes of each Gaussian's color also match with large changes of its position in terms of a ratio. To this end, we additionally introduce novel self-learning strategies to improve the learning of the neural view-dependent color field without any external knowledge or priors, aiming for better generalization on unseen scenes. Our numerical and visual comparisons on widely used benchmarks justify our idea and show better ability of high fidelity geometry recovery in 3DGS. Project page: \url{https://github.com/yangyixi21/Neural-Color-Fields-3DGS}.

\end{abstract}

\section{Introduction}
\vspace{-0.2cm}
3DGS~\cite{3dgs} has become a crucial representation for radiance fields due to its high efficiency and good quality in rendering. 3DGS employs discrete 3D Gaussians with additional attributes like color, scale, and opacity to approximate a continuous and smooth radiance field, which can be rendered into RGB, depth, and normal images. While the approximation does not make up the discrete essence of 3D Gaussians, which usually struggles to fit high-frequency color change, resulting in rough and incomplete geometry recovery.

Recent effort focuses on constraints that can directly get imposed on geometry. These methods usually employ various strategies such as continuous distance fields~\cite{gspull}, multi-view consistency~\cite{pgsr}, and monocular or geometry consistent priors~\cite{monosdf,zhang2025monoinstance}, leading to better surface inference. However, these strategies suffer from sensitive mappings from signed/unsigned distances to density, extensive verification of consistency across multiple views in iterative optimization, or limited generalization ability of external knowledge or priors, respectively. Thus, more general solutions are still expected for smoother radiance fields represented through 3DGS.

To overcome this obstacle, we instead impose constraints on 3D Gaussian color within a neural view-dependent color field that can be learned without a need of any external prior or data-driven strategy. During a splatting procedure, each Gaussian first gets a color estimation from an implicit neural network, and then use it along with other learnable attributes like scale and opacity for alpha blending. Our motivation is two fold, both of which significantly improve the geometry inference of 3DGS within a smoother color field. First, the inherent spectral bias of the neural network towards low-frequency color signals encourages the geometry modeling
to explain high-frequency color variations in images. Second, we impose a constraint to improve the Lipschitz continuity of this neural color field, which ensures that significant changes in a Gaussian's color must correspond to significant shifts in its position. To this end, we parameterize
a view-dependent color field using a neural network and introduce novel strategies to improve the learning of the neural view-dependent color field along with the learning of 3D Gaussians, without a need of external priors or data-driven strategies, leading to better generalization on various unseen scenes. We justify our idea on widely used benchmarks in numerical and visual evaluations, which shows our superiority over the latest 3DGS variations as well. Our contributions are listed below.

\begin{itemize}
\item We introduce 3DGS within a neural view-dependent color field and novel self-supervision to learn the color field along with the learning of Gaussians with 3DGS.
\item We introduce a novel perspective of improving geometry inference with 3DGS by imposing continuous constraints on color, which encourages high-frequency geometry recovery.
\item We report the state-of-the-art performance on the widely used benchmarks and show advantages over other 3DGS variations.
\end{itemize}

\vspace{-0.4cm}
\section{Related Work}
\vspace{-0.3cm}
\subsection{Neural Radiance Fields and 3D Gaussian Splatting}
\vspace{-0.2cm}
NeRF~\cite{nerf} achieves photorealistic view synthesis with an implicit MLP representation, but suffers from high training and rendering costs. Subsequent works therefore explored hybrid grid-based and point-based representations~\cite{fridovich2022plenoxels,InstanceNgp,pointnerf}. 
Recently, 3D Gaussian Splatting (3DGS)~\cite{3dgs} has emerged as an efficient explicit scene representation. It models scenes with anisotropic 3D Gaussians and uses spherical harmonics to capture view-dependent colors, enabling real-time rendering with high fidelity. However, 
3DGS often struggles within textureless regions and recover high-frequency geometry details, leading to incomplete geometry and noisy depth. Our method builds on 3DGS and improves its attribute representation for better geometry reconstruction.

\vspace{-0.2cm}
\subsection{Geometry Reconstruction and Regularization in 3DGS}
\vspace{-0.2cm}
Since the geometric inference through 3DGS~\cite{3dgs} is usually unsatisfactory, a significant body of work has focused on improving its geometric reconstruction capabilities, typically by introducing geometric constraints or modifying the representation. For example, 2DGS~\cite{2dgs} considers view-dependent effects as a source of noise and proposes to model the scene using 2D Gaussians embedded in 3D space. By flattening 3D ellipsoids into 2D planes, these primitives adhere more closely to object surfaces, achieving clearer surfaces and fewer floating artifacts. Similarly, PGSR~\cite{pgsr} also improves depth rendered by 3DGS~\cite{3dgs}. It proposes a novel geometry modeling with depth and enhances the smoothness and multi-view consistency by introducing various constraints. GSPrior~\cite{noda20263d} derives self-constrained geometric
priors from a TSDF grid fused from rendered depths.
For sparse-view surface reconstruction,
DP-GS~\cite{han2026sparse} propagates high-confidence depths
with normal priors, while VGGS~\cite{xiang2026vggs} uses
VGGT-derived geometric priors with anchor-calibrated depth
estimation and relative depth consistency. MaterialRefGS~\cite{zhang2025materialrefgs} uses multi-view material consistency for reflective scenes. Another line of work integrates geometric priors. For instance, SuGaR~\cite{sugar} regularizes the 3D Gaussians to align with the underlying surface and extracts a mesh from the Gaussian representation, thereby improving geometric consistency. Moreover, Anysplat~\cite{anysplat} learns priors to generate Gaussians.
\vspace{-0.2cm}
\subsection{Neural Implicit Fields and Regularization}
\vspace{-0.2cm}
Neural implicit fields~\cite{ma2021neuralpull} represent continuous signals with networks that favor low frequencies~\cite{rahaman2019spectralbiasneuralnetworks}. Frequency consolidation priors~\cite{chen2025sharpening} recover high-frequency geometry. While positional encoding~\cite{tancik2020fourierfeaturesletnetworks} typically counters this bias in NeRFs, we argue this inherent smoothness can serve as a geometric regularizer for explicit 3D Gaussian Splatting. NeRFPrior~\cite{zhang2025nerfprior} and VRP-UDF~\cite{zhang2026vrpudf} use radiance-field and volume-rendering priors for reconstruction, respectively; SparseRecon~\cite{han2025sparserecon} enforces feature and depth consistency. Moreover, recent methods like Splatfields~\cite{splatfields}, GOF~\cite{yu2024gof}, and NeRF-GS~\cite{fang2025nerf} utilize networks to model Gaussian attributes, while others~\cite{martins2024feature,qiu2024feature,zhou2024feature,li2024spacetime} incorporate high-dimensional feature fields. To improve stability, Lipschitz continuity prevents disproportionate output changes. For 3D geometry, constraining the Lipschitz constant aids learning from noisy data~\cite{liu2022learning}, with recent works constructing 1-Lipschitz neural distance fields for robust estimation~\cite{1lipschitz}. Level-set alignment~\cite{ma2023gradient} improves SDF gradient consistency. Crucially, Lipschitz constraints can link geometry and appearance. PermutoSDF~\cite{permutosdf} limits color field variation to force high-frequency signals into geometric structures. Distinct from these methods, we propose a direct approach that constrains the neural color field's Lipschitz constant, forcing large color variations to correspond to significant geometric changes and preventing overfitting to high-frequency noise.

\vspace{-1.0em}
\section{Methods}
\vspace{-0.2cm}
\begin{figure*}[t]
  \centering

  \includegraphics[width=1.0\linewidth]{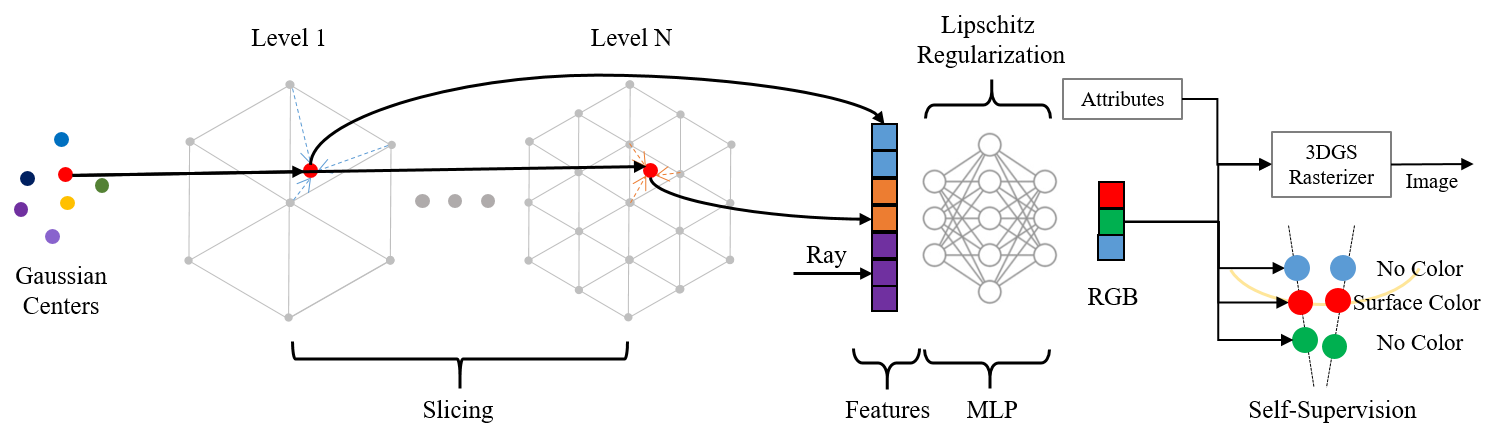}
  
  \caption{
  Overview of our framework. We first use Gaussian centers to query multi-resolution feature grids to obtain features. Then, together with view directions, we employ an MLP to predict view-dependent colors for each Gaussian. The predicted colors and Gaussian attributes are then rendered by the differentiable rasterizer. We impose self-learned supervision on the neural color field.
  }
  \vspace{-0.6cm}
  \label{fig:overview}
\end{figure*}
\subsection{Overview and Background}
\noindent\textbf{Overview. }As shown in Fig.~\ref{fig:overview}, we learn 3D Gaussians $\{\bm{G}_j\}_{j=1}^J$ (with center $\bm{m}_j$, opacity $o_j$, scale $s_j$, color $\bm{c}_j$) from RGB images $\{\bm{V}_i\}_{i=1}^I$ via 3DGS. We also learn a neural implicit function $F(\bm{f}_j,\bm{r})$ that maps a feature $\bm{f}_j$ and a ray direction $\bm{r}$ pointing to a Gaussian to a view-dependent color $\bm{c}_j^r$. During splatting, each Gaussian queries $F$ for its color before alpha blending. The color field is learned with rendering supervision and a novel self-supervision, leading to smoother radiance fields and better geometry.

\noindent\textbf{Rendering with 3DGS. }With camera pose $\bm{p}_i$, we render RGB $\bm{V}_i'$ and depth $\bm{D}_i'$ by splatting the Gaussians $\{\bm{G}_j\}_{j=1}^J$ below. We can learn 3D Gaussians by minimizing the discrepancy $L_{Img}$ in Eq.~\eqref{eq:Loss} between the rendered $\{\bm{V}_i',\bm{D}_i'\}$ and the observed $\{\bm{V}_i,\bm{D}_i\}$, where the depth rendering is optional,

\vspace{-4mm}
\begin{equation}
\label{eq:rendering}
\{\bm{V}_i',\bm{D}_i'\}=splatting (\{\bm{G}_j\},\bm{p}_i).
\end{equation}
\vspace{-8mm}

\subsection{Neural View-dependent Color Fields}
\vspace{-0.2cm}
We use an implicit function $F$ to represent the view-dependent color field using a neural network. For each Gaussian $\bm{G}_j$, we first employ a hash feature grid $\bm{H}$ to model the spatial feature $\bm{f}$ in 3D. We use its center $\bm{m}_j$ to interpolate a feature $\bm{f}_j$ from the grid $\bm{H}$, which is then used to predict a color $\bm{c}_j^r$ through the neural network, where $\bm{r}$ is the normalized direction from the camera center to the Gaussian center $\bm{m}_j$. The MLP is queried once per camera--Gaussian pair, and
the predicted color is shared across its projected footprint,

\vspace{-4mm}
\begin{equation}
\label{eq:rendering_color}
\bm{c}_j^r = F(\bm{f}_j,\bm{r}), \quad \text{where} \quad \bm{f}_j = \bm{H} (\bm{m}_j).
\end{equation}

To improve the efficiency of feature interpolation from $\bm{H}$, we do not follow INGP~\cite{InstanceNgp} which uses a cubical grid and linearly interpolate features from the hashed eight corners of the containing cell. Instead, we adopt a tetrahedral lattice~\cite{permutosdf,learningdeformabletetrahedralmeshes} which only needs four corners in each cell in interpolation. It reduces both of the storage and the computation complexity by a factor of two. 

\vspace{-0.2cm}
\subsection{Self-Learning of Neural View-dependent Color Field}
\label{sec:mlp_method}
\vspace{-0.2cm}

We learn the hash feature lattice $\bm{H}$ and the implicit function $F$ from the radiance field represented by current Gaussians $\{\bm{G}_j\}$ in iterative optimization. We use four kinds of supervision signals, one is from the 2D rendering error to the observation, the other three are from the 3D space. It is intuitive to use the 2D rendering error to update both of $\bm{H}$ and $\bm{F}$, since the whole framework is differentiable based on 3DGS, thus, we will focus on introducing the 3D supervision below.

\begin{wrapfigure}{r}{0.5\linewidth}
  \centering
  \vspace{-10pt}
  
  \includegraphics[width=\linewidth]{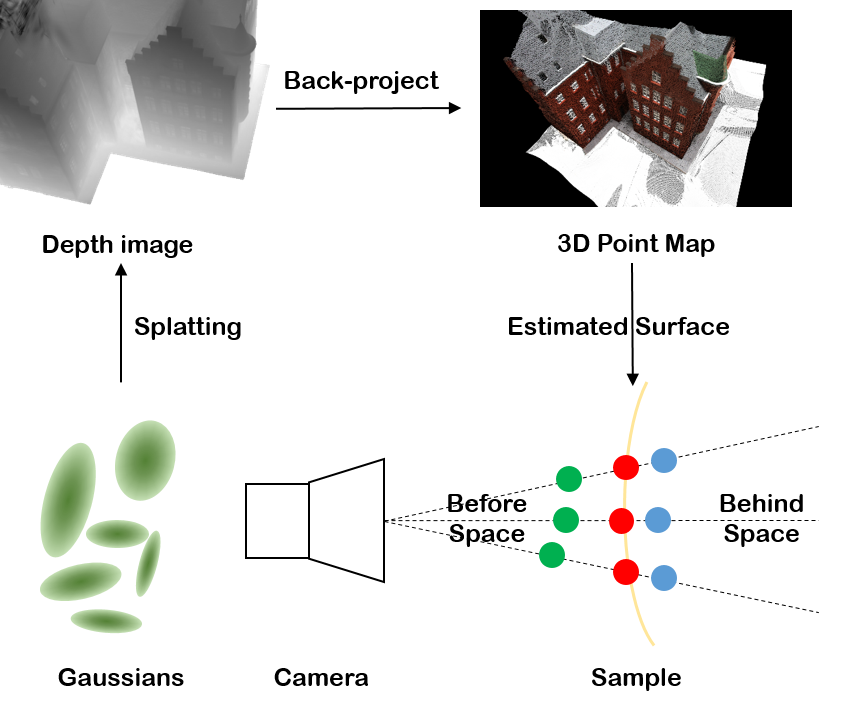}
  
  \caption{
Illustration of our 3D supervision sampling strategy.
Given the rendered depth $\bm{D}_i'$, we back-project each pixel to obtain the surface intersection.
(a) On-surface: the intersection point ($\pi^{-1}(\bm{D}_i')$, red) is supervised by the ground-truth color in $\bm{V}_i$ (Eq.~\eqref{eq:onsurface}).
(b) Free space: points ${\bm{e}}_i$ sampled along the ray around the intersection are constrained to have no color before (${\bm{e}}_i^{-}$, green) and behind (${\bm{e}}_i^{+}$, blue) the surface (Eq.~\eqref{eq:beforeandaftersurface}).
}
  \label{fig:sampling_strategy}
  
  \vspace{-10pt}
\end{wrapfigure}

\noindent\textbf{On Surface Sampling. }As illustrated in Fig.~\ref{fig:sampling_strategy}, we produce samples on the estimated surface to learn the view-dependent color field. We first render a depth map $\bm{D}_i'$ with a camera pose $\bm{p}_i$, and regard the 3D points $\pi^{-1}(\bm{D}_i')$ back-projected from $\bm{D}_i'$ as the training samples, where $\pi^{-1}$ denotes the back projection. We collect the coordinates $\pi^{-1}(\bm{D}_i')$ of these points, the ray direction from the center of the $i$-th camera to each 3D Gaussian, and the color on the image $\bm{V}_i$ as the supervision. The rationale behind this supervision is that the color we see on images corresponds to the color on the surface. Then, we minimize the prediction error to the color on $\bm{V}_i$,

\vspace{-4mm}
\begin{equation}
\label{eq:onsurface}
L_{On} = ||F(\bm{H}(\pi^{-1}(\bm{D}_i')),\pi^{-1}(\bm{D}_i')-\bm{p}_i^t) - \bm{V}_i||_2^2,
\end{equation}

\noindent where $\bm{p}_i^t$ is the center of the $i$-th camera.

\noindent\textbf{Free Space Sampling before and behind the Surface. }
Given the estimated surface intersections $\pi^{-1}(\bm{D}_i')$, we assume that the space before and slightly behind each intersection should be empty. We therefore sample points along each ray around the intersection, as shown in Fig.~\ref{fig:sampling_strategy}, and constrain their colors to be zero. Specifically, we sample ${\bm{e}}_i \sim \mathcal{N}(\bm{D}_i', \delta^2)$ and split the samples into points before the surface ${\bm{e}}_i^{-}$ and behind the surface ${\bm{e}}_i^{+}$, which are weighted with different confidence,

\vspace{-4mm}
\begin{equation}
\label{eq:beforeandaftersurface}
L_{Before}=\|F(\bm{H}({\bm{e}}_i^{-}),{\bm{e}}_i^{-}-\bm{p}_i^t)\|_2^2,\quad
L_{Behind}=\|F(\bm{H}({\bm{e}}_i^{+}),{\bm{e}}_i^{+}-\bm{p}_i^t)\|_2^2.
\end{equation}

\noindent\textbf{Lipschitz Continuity of the color Field. }
We impose a Lipschitz constraint on $F$ to encourage smoothness in the neural color field and prevent high-frequency color variation from being explained by appearance alone. This encourages large color changes to align with geometric variation. Formally, $F$ is $k$-Lipschitz continuous if,

\vspace{-4mm}
\begin{equation}
\label{eq:lipschitz}
|F(\bm{H}(\bm{m}))-F(\bm{H}(\bm{m}'))| \le k|\bm{m}-\bm{m}'|,
\end{equation}

\noindent where we omit the ray direction $\bm{r}$ for simplicity, and $k$ bounds the rate of change of $F$.
We employ recent work~\cite{liu2022learning} to softly add the $k$-Lipschitz constraints through the weights of neural networks. For a given MLP layer $y=g(\bm{W}_ux+b_u)$, we employ a learnable Lipschitz bound $k_u$ for each layer to replace the original weight $\bm{W}_u$ with $\overline{\bm{W}_u}$,

\vspace{-4mm}
\begin{equation}
\label{eq:weight}
\overline{\bm{W}_u} = normalization (\bm{W}_u,softplus(k_u)),
\end{equation}

\noindent where $softplus(k_u)=ln(1+e^{k_u})$ and the normalization scales each row of $\bm{W}_u$ such that the absolute value of the row-sum is less than or equals to $softplus(k_u)$. The Lipschitz bound for the whole network is the product of the per-layer Lipschitz constraint, which is what we regularize over all layers,

\vspace{-4mm}
\begin{equation}
\label{eq:LipConstraints}
L_{Lip}=\prod_{u=1}^U softplus(k_u).
\end{equation}
\vspace{-4mm}

\subsection{Loss and Details}
\vspace{-0.2cm}

\noindent\textbf{Loss. }We conduct the optimization through 3DGS to learn spatial feature lattice $\bm{H}$, the function $F$, and the Gaussians $\{\bm{G}_j\}$ from RGB images,

\begin{equation}
\label{eq:Loss}
\min_{\bm{H},F,\{\bm{G}_j\}} L_{Img} + \alpha L_{On} + \beta L_{Before} + \gamma L_{Behind} + \eta L_{Lip},
\end{equation}

\noindent where $L_{Img}$ is the rendering error from 3DGS, and $\alpha$, $\beta$, $\gamma$, and $\eta$ are balance weights. We impose different weights on sampling before and behind the estimated surface since we got higher confidence that there is no points before the surface while there is uncertainty behind the estimated surface.

\noindent\textbf{Implementation Details. }
To ensure a stable and efficient training, we adopt three key strategies. 
First, we utilize a 3D bounding box to mark the near front. The proposed neural color field (Hash Grid + MLP) is applied exclusively to Gaussians within this box to capture fine details, while background Gaussians retain standard Spherical Harmonics. 
Second, we employ gradient detachment to decouple appearance from geometry, detaching gradients for Gaussian attributes (e.g., position $\bm{m}$) during 3D supervision ($L_{On, Before, Behind}$) so these losses only update the feature grid $\bm{H}$ and implicit network $F$. 
Third, we use the asymmetric sampling for free-space supervision, applying distinct variances for points sampled along the ray before and behind the surface.

Regarding the network architecture, we use a 4-layer MLP for color decoding. We set the hash grid to 18 levels with 32 feature dimensions for DTU~\cite{aanaes2016large} and increase this to 20 levels with 48 dimensions for the TnT~\cite{knapitsch2017tanks} and Mip-NeRF~\cite{barron2022mipnerf360unboundedantialiased} 360 datasets. Sampling variances are set to $\delta_{before}=0.1, \delta_{behind}=0.01$ for DTU, and increased to $1.0$ and $0.1$ for other datasets. Loss weights $\alpha, \beta, \gamma$ in Eq.~\eqref{eq:Loss} are $0.05$ for TnT and $0.01$ otherwise. Lipschitz constraints regarding Eq.~\eqref{eq:LipConstraints} with $\eta=10^{-6}$ are imposed after 10k iterations within a total of 30k. The Geometry is extracted via TSDF fusion (voxel size 0.004) and the marching cubes algorithm using Open3D~\cite{open3d}. On Mip-NeRF 360, we adhere to the standard protocol by holding out $1/8$ of images for testing.
\vspace{-0.4cm}
\section{Experiments}
\vspace{-0.2cm}
\subsection{Experimental Setup}
\vspace{-0.2cm}
\noindent\textbf{Datasets.}
To comprehensively evaluate the effectiveness of our method, we conduct extensive experiments across three standard benchmarks covering both object-level details and large-scale real-world scenes including DTU~\cite{aanaes2016large}, TnT~\cite{knapitsch2017tanks}, and Mip-NeRF360~\cite{barron2022mipnerf360unboundedantialiased}.

\vspace{-0.4em}
\noindent\textbf{Baselines.}
We compare our method with state-of-the-art 3DGS variants, including 3DGS~\cite{3dgs}, 2DGS~\cite{2dgs}, PGSR~\cite{pgsr}, and FastGS~\cite{fastgs}, as well as SplatFields~\cite{splatfields}, which also uses MLP-predicted attributes. For real-world scenes without ground-truth meshes, we further compare rendering quality with Instant-NGP~\cite{InstanceNgp}, Mip-NeRF 360~\cite{barron2022mipnerf360unboundedantialiased}, and BakedSDF~\cite{baskedSDF}.  We conduct all the experiments on a single NVIDIA GeForce RTX 3090.

\vspace{-0.4em}
\noindent\textbf{Evaluation Metrics.}
We evaluate the geometry using Chamfer Distance (CD) on DTU and F-score on TnT, following the official protocols and previous methods.
For novel view synthesis on Mip-NeRF 360, we report PSNR, SSIM, and LPIPS, which are widely used metrics evaluating rendering quality.

\begin{table}[h]
\vspace{-0.2cm}
  \centering
    \caption{CD comparison on DTU. The top 3 are ranked as \colorbox{red}{1st}, \colorbox{orange}{2nd} and \colorbox{yellow}{3rd}.
    }
  \vspace{-0.2cm}
  \label{tab:dtu}
  \begin{adjustbox}{width=\textwidth,center}
  \begin{tabular}{l|ccccccccccccccc|c|c}
    \toprule
     Methods & 24 & 37 & 40 & 55 & 63 & 65 & 69 & 83 & 97 & 105 & 106 & 110 & 114 & 118 & 122 & Mean & Time  \\
    \midrule
     
     SplatFields~\cite{splatfields}& 3.76 & 4.20 & 4.35 & 2.16 & 4.31 & 5.93 & 3.03 & 3.03 & 3.44 & 2.49 & 5.31 & 4.94 & 4.19 & 3.33 & 2.85 & 3.82 & 1h\\
     Mini-Splatting~\cite{minisplatting} & 2.05 & 1.24 & 1.55 & 1.11 & 2.14 & 1.82 & 1.59 & 2.12 & 1.87 & 1.57 & 1.48 & 2.20 & 1.78 & 1.29 & 1.33 & 1.67 & \cellcolor{yellow}6.5m \\  
     AnySplat~\cite{anysplat} & 2.93 & 2.95 & 2.15 & 2.65 & 4.09 & 2.68 & 1.94 & 2.85 & 2.25 & 2.55 & 2.24 & 4.30 & 2.21 & 2.00 & 2.37 & 2.68 & \cellcolor{red}3m\\
     FastGS~\cite{fastgs} & 2.66 & 2.04 & 2.45 & 2.22 & 3.73 & 2.62 & 1.81 & 2.53 & 2.21 & 2.30 & 2.54 & 2.38 & 2.81 & 1.54 & 1.67 & 2.37 & \cellcolor{orange}3.1m \\  
     3DGS~\cite{3dgs} & 1.97 & 1.37 & 2.11 & 1.73 & 3.52 & 2.56 & 1.67 & 1.93 & 1.75 & 1.84 & 1.72 & 2.24 & 1.57 & 1.43 & 1.33 & 1.92 &  22.3m  \\
     2DGS~\cite{2dgs} & \cellcolor{yellow}0.48 & 0.91 & \cellcolor{orange}0.39 & \cellcolor{yellow}0.39 & \cellcolor{orange}1.01 & 0.83 & \cellcolor{yellow}0.81 & 1.36 & 1.27 & 0.76 & \cellcolor{orange}0.70 & 1.40 & \cellcolor{yellow}0.40 & 0.76 & \cellcolor{yellow}0.52 & 0.80 & 20.5m   \\
     PGSR~\cite{pgsr} & \cellcolor{orange}0.37 & \cellcolor{red}0.55 & \cellcolor{orange}0.39 & \cellcolor{orange}0.36 & \cellcolor{red}0.78 & \cellcolor{orange}0.58 & \cellcolor{red}0.49 & \cellcolor{orange}1.10 & \cellcolor{red}0.64 & \cellcolor{orange}0.59 & \cellcolor{red}0.47 & \cellcolor{orange}0.56 & \cellcolor{orange}0.32 & \cellcolor{red}0.37 & \cellcolor{red}0.35 & \cellcolor{orange}0.53 & 32.5m \\
     \midrule
     Ours(FastGS) & 1.43 & 1.14 & 1.75 & 1.15 & 2.59 & 1.37 & 1.67 & 2.38 & 1.80 & 1.05 & 1.84 & 2.61 & 0.98 & 1.54 & 1.60 & 1.66 & 12.3m \\ 
     Ours(3DGS) & 0.82 & \cellcolor{yellow}0.88 & 1.64 & 1.05 & 1.94 & 1.00 & 1.18 & 1.21 & 1.29 & 0.73 & 0.93 & 2.07 & 0.64 & 0.82 & 0.70 & 1.13 &  30.1m \\
     Ours(2DGS) & 0.55 & \cellcolor{orange}0.73 & \cellcolor{yellow}0.43 & 0.43 & \cellcolor{yellow}1.24 & \cellcolor{yellow}0.81 & 0.86 & \cellcolor{yellow}1.12 & \cellcolor{yellow}1.23 & \cellcolor{yellow}0.64 & \cellcolor{yellow}0.80 & \cellcolor{yellow}1.30 & \cellcolor{yellow}0.40 & \cellcolor{yellow}0.59 & \cellcolor{orange}0.48 & \cellcolor{yellow}0.77 &  25.1m \\
     Ours(PGSR) & \cellcolor{red}0.33 & \cellcolor{red}0.55 & \cellcolor{red}0.38 & \cellcolor{red}0.34 & \cellcolor{red}0.78 & \cellcolor{red}0.50 & \cellcolor{orange}0.50 & \cellcolor{red}1.05 & \cellcolor{orange}0.66 & \cellcolor{red}0.58 & \cellcolor{red}0.47 & \cellcolor{red}0.50 & \cellcolor{red}0.31 & \cellcolor{orange}0.38 & \cellcolor{red}0.35 & \cellcolor{red}0.51 & 42.4m \\
  \bottomrule
    \end{tabular}
    \end{adjustbox}
\vspace{-0.6cm}
\end{table}

\subsection{Evaluations}
\begin{figure*}[t]   %
  \centering         %

  \centerline{\includegraphics[width=0.9\linewidth]{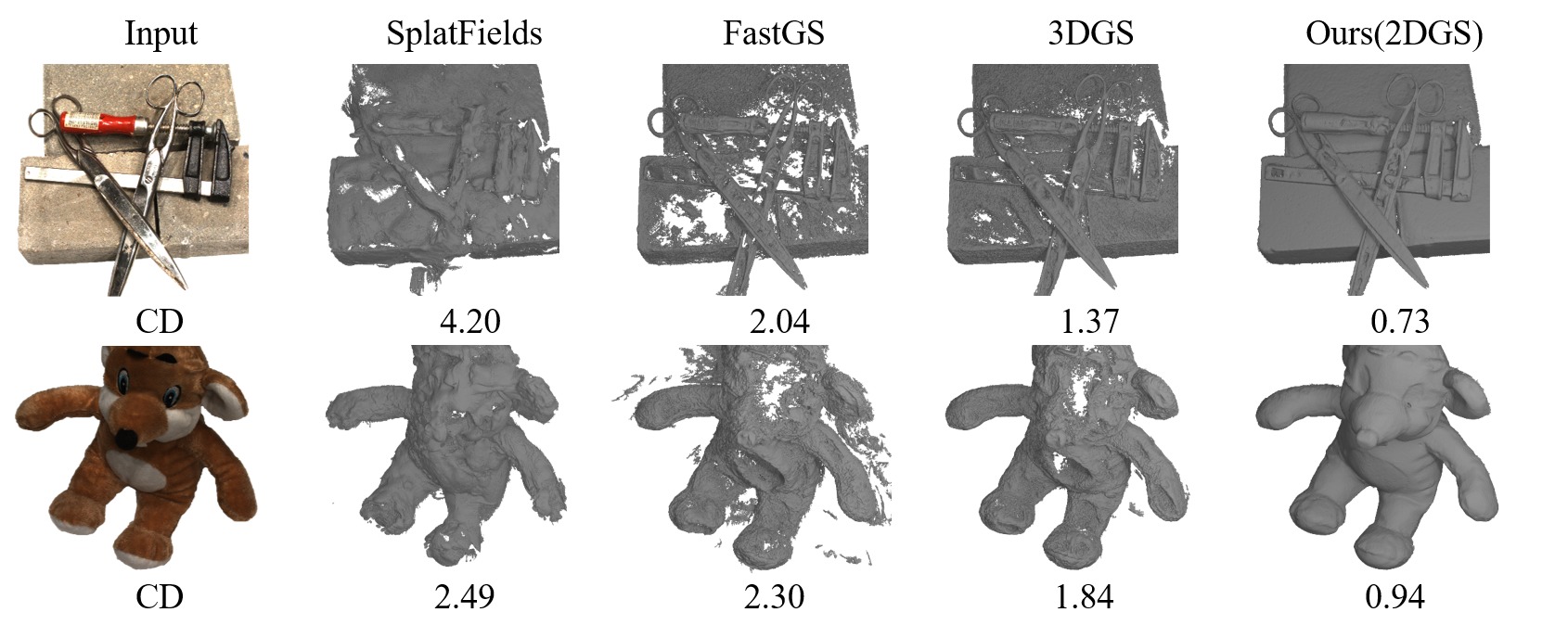}}
  \centerline{\small (a) Comparison with State-of-the-Art Methods.}
  
  \vspace{0.3cm}
  
  \centerline{\includegraphics[width=0.9\linewidth]{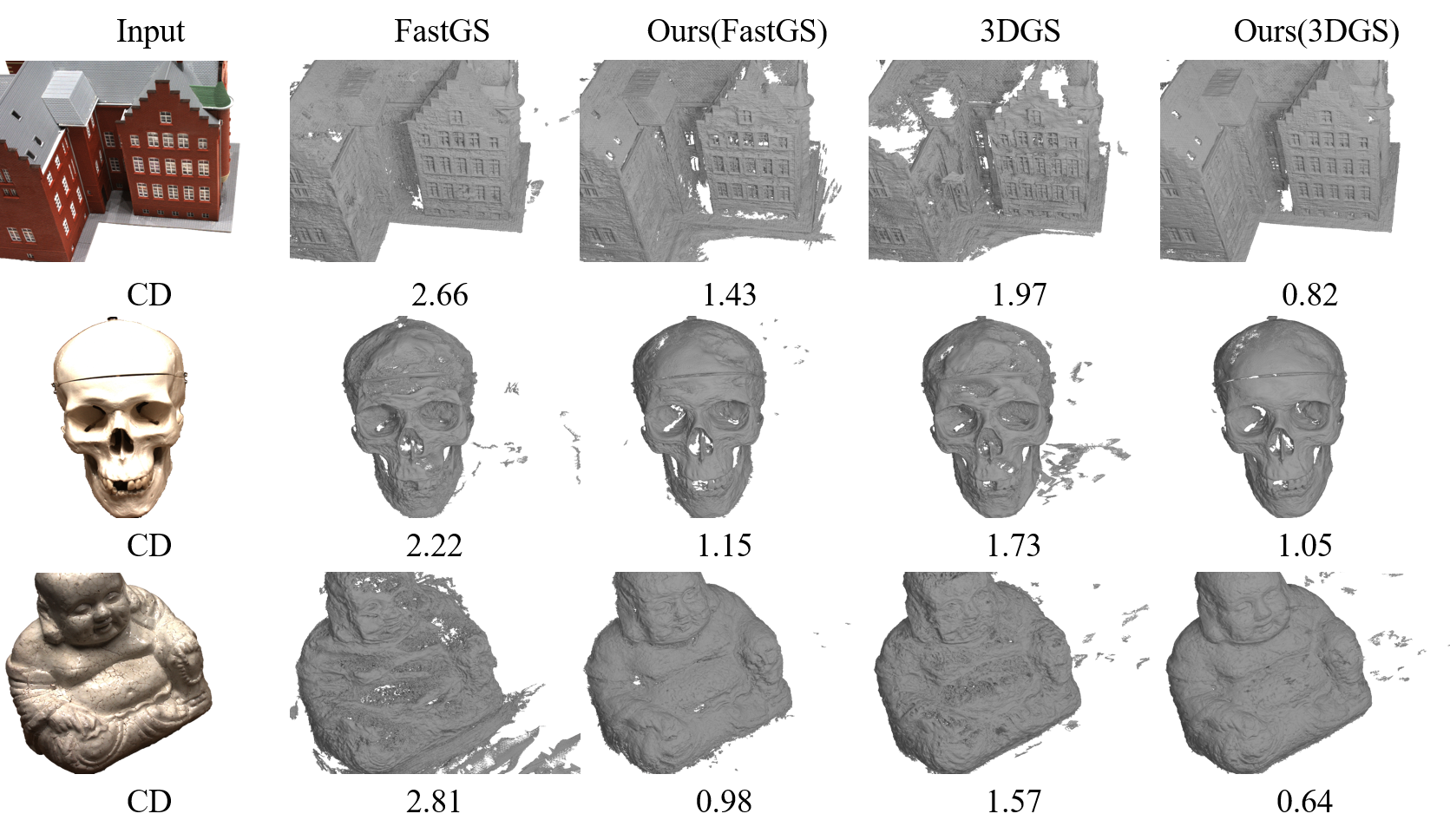}}
  \centerline{\small (b) Improvements on Various Backbones.}

  \caption{
    Performance on the DTU dataset.
    (a) Comparison of Ours (2DGS) with existing Gaussian Splatting methods.
    (b) Improvements over FastGS and 3DGS backbones.
  }
  \label{fig:dtu_analysis_vertical}
\end{figure*}

\vspace{-0.2cm}
\begin{table}[!t]

  \centering
    \caption{Comparisons of rendering quality on DTU. The top 3 are ranked as \colorbox{red}{1st}, \colorbox{orange}{2nd} and \colorbox{yellow}{3rd}.}
  \vspace{0.0cm}
  \label{tab:dtu_render}
  \begin{tabular}{l|ccc}
    \toprule

      Method & PSNR$\uparrow$ & SSIM$\uparrow$ & LPIPS$\downarrow$\\
    \midrule
    SplatFields~\cite{splatfields}&24.14&0.813&0.241\\
    Mini-Splatting~\cite{minisplatting} & 30.39 & 0.946& 0.071\\
    FastGS~\cite{fastgs} & 30.84 &  0.950 & 0.073\\
    
     3DGS~\cite{3dgs} &   \cellcolor{yellow}32.56 &  0.953 & \cellcolor{yellow}0.063 \\
     2DGS~\cite{2dgs} & 32.32 & \cellcolor{yellow}0.955 & \cellcolor{red}0.058 \\
     \midrule
     Ours(FastGS) & 31.06 &  0.951 & 0.070 \\
     Ours(3DGS) & \cellcolor{red}32.77 &  \cellcolor{red}0.956 & 0.065\\
     Ours(2DGS) & \cellcolor{orange}32.61 &  \cellcolor{orange}0.955 &\cellcolor{orange}0.060\\

  \bottomrule
    \end{tabular}
\vspace{-0.4cm}
\end{table}
\noindent\textbf{Comparisons on DTU. }We first evaluate our method by comparing it with other 3DGS variations on DTU, as reported in Tab.~\ref{tab:dtu}. To evaluate our general capability of working with various 3DGS variations, we report our results with vanilla 3DGS~\cite{3dgs}, 2DGS~\cite{2dgs}, FastGS~\cite{fastgs}, and PGSR~\cite{pgsr}. We include vanilla 3DGS to highlight our performance without additional losses for geometry inference since the additional constraints used by 2DGS~\cite{2dgs}, FastGS~\cite{fastgs}, and PGSR~\cite{pgsr} are redundant to what we can improve, which weakens our improvements. The comparison shows that our method improves the average
reconstruction accuracy of all four backbones. Specifically, our method improves the performance by $30.0\%$ over FastGS, $41.1\%$ over 3DGS, and $3.8\%$ over 2DGS in average, which can be also observed in visual comparison in Fig.~\ref{fig:dtu_analysis_vertical} (a). For PGSR, our method further reduces the average CD from 0.53 to 0.51. PGSR already achieves strong reconstruction accuracy through geometry-oriented constraints, which may partially overlap with the effects of our regularization and leave less room for further performance improvements on DTU. Nevertheless, this additional gain demonstrates that our method can complement an already strong geometry-constrained backbone.
\begin{figure}[h]
\vspace{-0.2cm}
  \centering
  \includegraphics[width=1.0\linewidth]{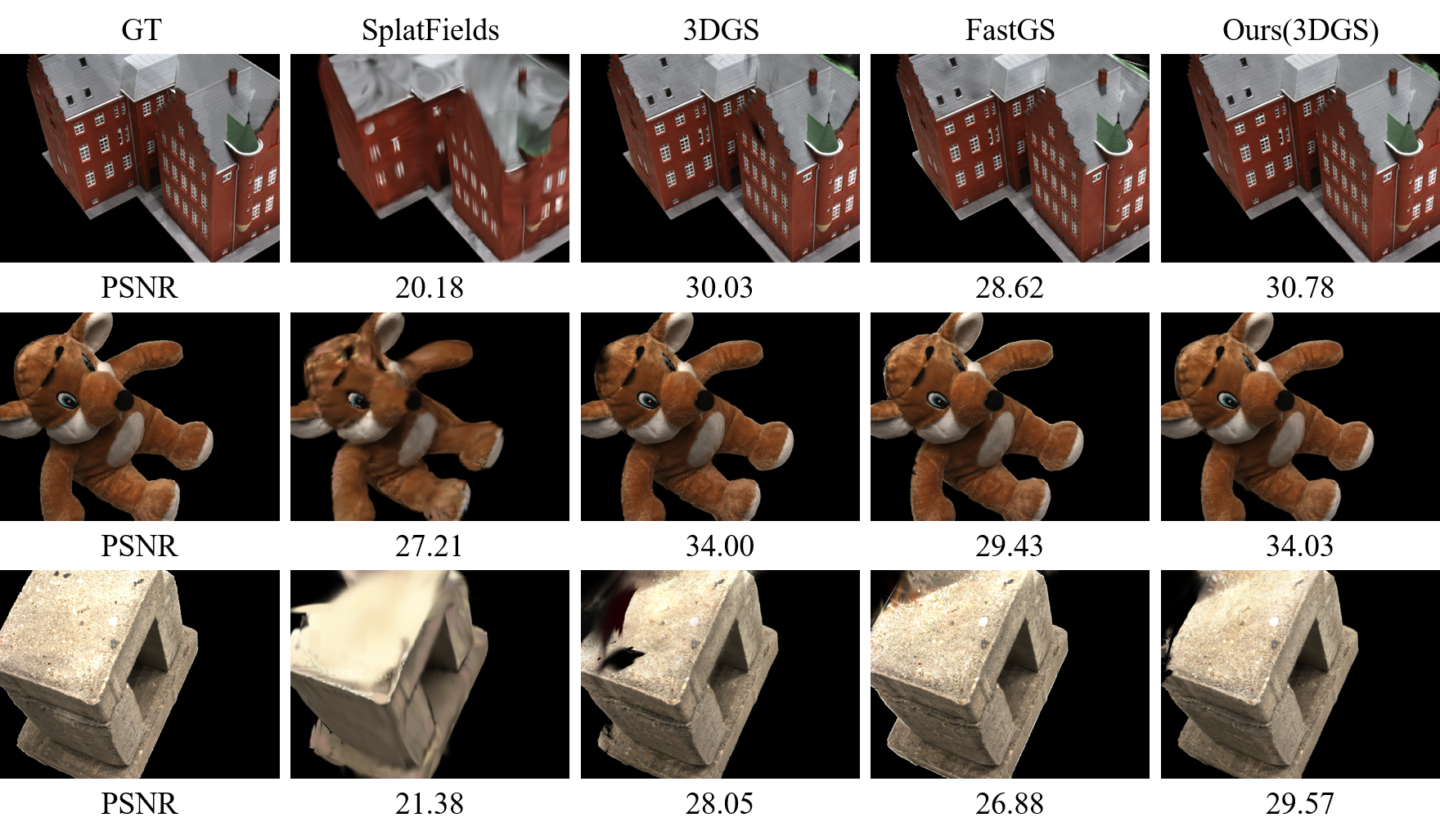}
  \vspace{-0.6cm}
  \caption{Visual comparison of rendering on DTU dataset.}
  \label{fig:rendering_comparison}
\vspace{-0.4cm}
\end{figure}

Moreover, our geometry inference also outperform other 3DGS variations like Mini-Splatting~\cite{minisplatting}, SplatFields~\cite{splatfields}, and Anysplat~\cite{anysplat}. Compared to Splatfields and Anysplat which also leverage an MLP to predict the Gaussian attributes in either overfitting or data-driven manner, our methods merely predict color, and only add constraints on color, which can push attributes related to geometry to match high-frequency color change. To highlight the improvement, we show visual comparison in Fig.~\ref{fig:dtu_analysis_vertical} (a), where our method recovers more complete and compact surfaces.

\begin{wraptable}{r}{0.52\textwidth}
  \vspace{-0.4cm}
  \centering
  \caption{Comparisons of rendering quality on TnT. The top 3 are ranked as \colorbox{red}{1st}, \colorbox{orange}{2nd} and \colorbox{yellow}{3rd}.}
  \vspace{-0.0cm}
  \label{tab:tnt_render}
  \begin{tabular}{l|ccc}
    \toprule

      Method & PSNR$\uparrow$ & SSIM$\uparrow$ & LPIPS$\downarrow$\\
    \midrule
    SplatFields~\cite{splatfields}& 15.99 & 0.518 & 0.650 \\

    FastGS~\cite{fastgs} & \cellcolor{red}25.67  &  \cellcolor{orange}0.841  & \cellcolor{orange}0.241\\
    
     3DGS~\cite{3dgs} & 24.37  & 0.812   & 0.265  \\
    \midrule
     Ours(FastGS) &  \cellcolor{orange}25.13 &   \cellcolor{yellow}0.834  & \cellcolor{red}0.225 \\
     Ours(3DGS) & \cellcolor{yellow}24.76  & \cellcolor{red}0.849   &  \cellcolor{yellow}0.245 \\

  \bottomrule
    \end{tabular}
\vspace{-0.4cm}
\end{wraptable}
Tab.~\ref{tab:dtu_render} and Fig.~\ref{fig:rendering_comparison} also report our comparable performance in novel view synthesis. Although we use neural networks to model the color field, the bias of neural networks on low-frequency color change does not hurt our appearance modeling in radiance fields, since our method can encourage larger geometry change to make up the weakness of low-frequency color change with the Lipschitz constraint. For the efficiency, the use of neural network slightly slows down the training.

\begin{table}[h!]
\vspace{-0.1cm}
  \centering
  \vspace{-0.2cm}
  \caption{F-score comparisons on Tanks And Temples dataset~\cite{knapitsch2017tanks}. }
  \vspace{-0.2cm}
  \label{tab:tnt}
  \begin{adjustbox}{width=\textwidth,center}
  \begin{tabular}{l|cccccc|c}
    \toprule
     Methods & Barn & Caterpillar & Courthouse & Ignatius & Meetingroom & Truck & Mean   \\
    \midrule
    SplatFields~\cite{splatfields}& 0.02 & 0.03 &0.01 & 0.05 &0.01 & 0.05&0.03 \\
     FastGS~\cite{fastgs} & 0.10 & 0.09 & 0.07 & 0.09 & 0.02 & 0.12  &  0.08\\
     3DGS~\cite{3dgs} & 0.13 & 0.08 & 0.09 & 0.25 & 0.06 & 0.18 & 0.13 \\
         \midrule
     Ours(FastGS) & 0.11 & 0.10 & 0.07 & 0.14 & 0.04 & 0.16 & 0.11\\
     
     Ours(3DGS) & 0.17 & 0.14 & 0.10 & 0.28 & 0.09 & 0.19 & 0.16\\
     
  \bottomrule
    \end{tabular}
    \end{adjustbox}
\vspace{-0.8cm}
\end{table}

\begin{figure*}[t]
  \centering

  \includegraphics[width=1.0\linewidth]{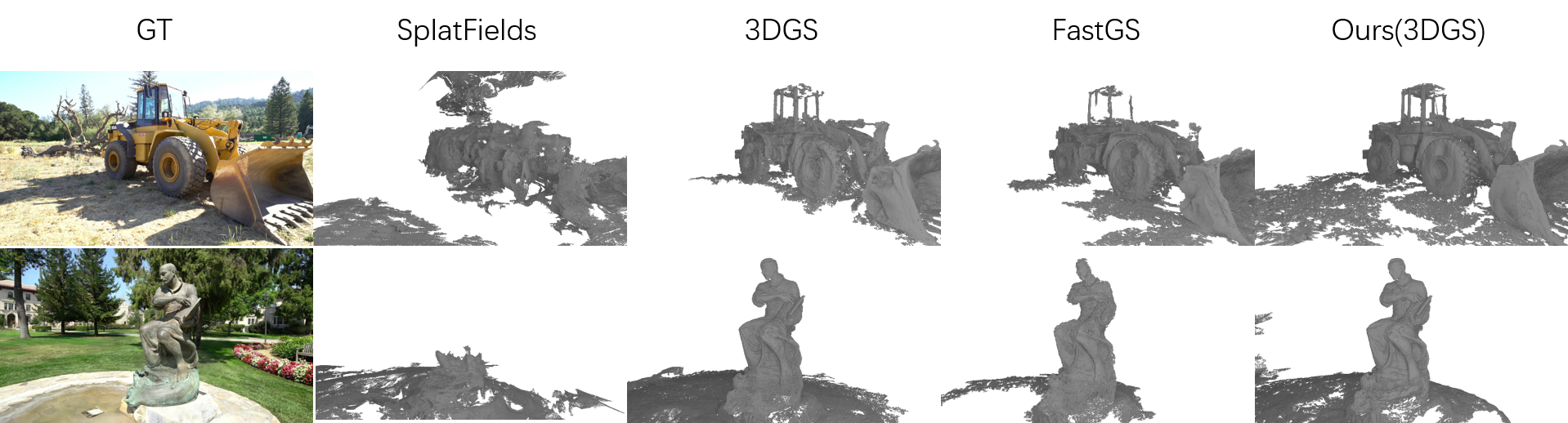}
  
  \caption{
  Visual comparisons of geometry inference on TnT~\cite{knapitsch2017tanks}.
  }
  \label{fig:tnt_compare_visual}
\vspace{-0.4cm}
\end{figure*}

\noindent\textbf{Comparisons on TnT. }
Tab.~\ref{tab:tnt} shows results on the large-scale TnT dataset with FastGS and 3DGS backbones. Our method improves F-score for both variants, demonstrating better geometry reconstruction on realistic large-scale scenes, similar to our improvements shown in Fig.~\ref{fig:tnt_compare_visual}. Although SplatFields is also MLP-based, it does not scale well to large scenes. In contrast, our spatial feature grid and neural color field generalize better under challenging indoor and outdoor lighting.

\begin{table}
  
  \centering
  \caption{Comparisons of rendering quality on MipNeRF-360.  }
  \vspace{-0.0cm}
  \label{tab:mipnerf360}
  \resizebox{0.7\textwidth}{!}{\begin{tabular}{l|ccc|ccc}
    \toprule
       & \multicolumn{3}{c|}{Indoor Scene} & \multicolumn{3}{c}{Outdoor Scene} \\
    \midrule
       & PSNR$\uparrow$ & SSIM$\uparrow$ & LPIPS$\downarrow$ & PSNR$\uparrow$ & SSIM$\uparrow$ & LPIPS$\downarrow$  \\
    \cmidrule{1-7}
     NeRF\cite{nerf}        & 26.84 & 0.790 & 0.370 & 21.46 & 0.458 & 0.515     \\
     Instant-NGP~\cite{InstanceNgp} & 29.15 & 0.880 & 0.216 & 22.90 & 0.566 & 0.371  \\
     MipNeRF 360~\cite{barron2022mipnerf360unboundedantialiased} &  \cellcolor{red}31.72 & 0.917 &  \cellcolor{red}0.180 &  \cellcolor{yellow}24.47 & 0.691 &  \cellcolor{yellow}0.283 \\
     BakedSDF~\cite{baskedSDF} & 27.06 & 0.839 & 0.258 & 22.47 & 0.585 & 0.349 \\
    \cmidrule{1-7}
    SplatFields~\cite{splatfields}& 17.07 & 0.528 & 0.189 & 18.44 & 0.361 & 0.694 \\

    FastGS~\cite{fastgs} &  \cellcolor{orange}31.28 &  \cellcolor{orange}0.924 & \cellcolor{orange}0.184 &  \cellcolor{orange}24.53 &  \cellcolor{yellow}0.695 &  0.322 \\
    
     3DGS~\cite{3dgs} &  30.41 &  0.920 & 0.189 &  \cellcolor{red}24.64 &  \cellcolor{red}0.731 &  \cellcolor{red}0.272 \\
         \midrule
     Ours(FastGS) &  \cellcolor{yellow}30.67 &  \cellcolor{yellow}0.921 & \cellcolor{yellow}0.189 &  24.35 &  0.686 &  0.332 \\
     Ours(3DGS) & \cellcolor{yellow}30.43 &  \cellcolor{red}0.926 & 0.199 &  24.44 &  \cellcolor{orange}0.723 &  \cellcolor{orange}0.283 \\

  \bottomrule
    \end{tabular}}
\vspace{-0.2cm}
\end{table}

\begin{figure*}[!h]
  \centering

  \includegraphics[width=1.0\linewidth]{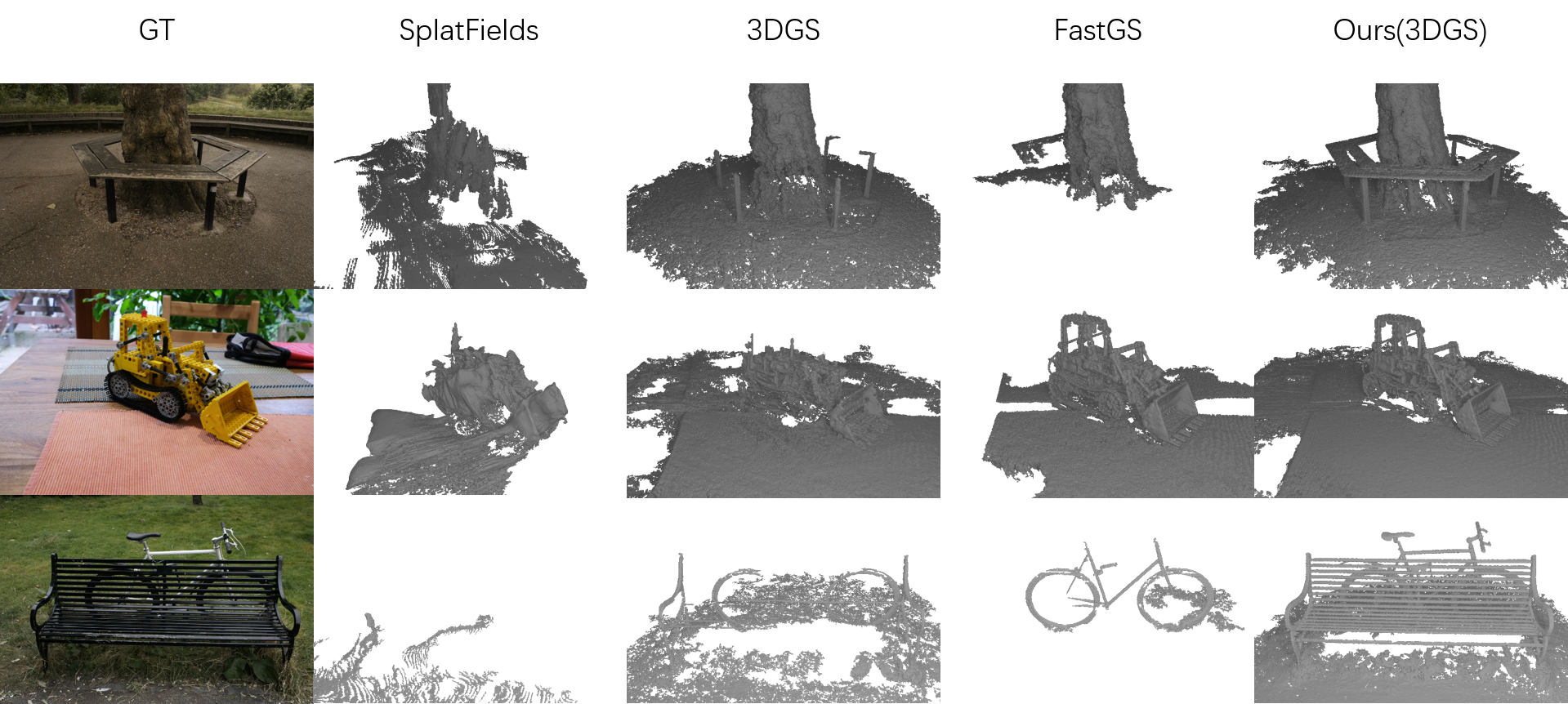}
  
  \caption{
  Visual comparisons between our method on MipNeRF-360~\cite{barron2022mipnerf360unboundedantialiased}.
  }
  \label{fig:mip_compare_visual}
\vspace{-0.4cm}
\end{figure*}

\noindent\textbf{Comparisons on MipNeRF-360. }We report our performance in novel view synthesis on MipNeRF-360 in Tab.~\ref{tab:mipnerf360}. Our comparable performance indicates that our method can promptly encourage geometry change to match with the large color change since we used a neural network that is biased on low-frequency signals to model the color field. Moreover, we also show our advantage with more complete surfaces and geometry details in geometry inference in Fig.~\ref{fig:mip_compare_visual}.

\vspace{-0.4cm}
\subsection{Ablation Study and Analysis}
\vspace{-0.2cm}
In this section, we conduct ablation studies on the key components of our framework to validate their effectiveness. Specifically, we investigate the contribution of (1) the Lipschitz continuity constraint on the color field, (2) the sampling strategy (both on surface sampling and free space sampling), and (3) the justification of our design.

\begin{table}[!h]
\vspace{-0.2cm}
\centering
\begin{minipage}{0.48\linewidth}
  \centering
  \caption{Ablation Study on DTU. Impact of different modules on geometry extraction (CD $\downarrow$).}
  \label{tab:ablation_horizontal}
  \resizebox{\linewidth}{!}{
      \begin{tabular}{l|cccc}
      \toprule
      Method & Base & w/ $L_{Lip}$  & w/ Samp.  & Full \\
      \midrule
      CD & 1.29 & 1.23 & 1.17 & \textbf{1.13} \\
      \bottomrule
      \end{tabular}
  }
\end{minipage}
\hfill %
\begin{minipage}{0.48\linewidth}
  \centering
  \caption{View-Dependence Analysis.}
  \label{tab:view_dependence}
  \resizebox{\linewidth}{!}{
      \begin{tabular}{l|cccc}
      \toprule
      Mode & CD$\downarrow$&PSNR $\uparrow$ & SSIM $\uparrow$ & LPIPS $\downarrow$ \\
      \midrule
      Non-View & 1.86&31.40 & 0.946 & 0.076 \\
      View-Dependent &\textbf{1.13}& \textbf{32.77} & \textbf{0.956} & \textbf{0.065} \\
      \bottomrule
      \end{tabular}
  }
\end{minipage}
\vspace{-0.2cm} %
\end{table}

\begin{figure}[t]
  \centering

  \includegraphics[width=0.95\linewidth]{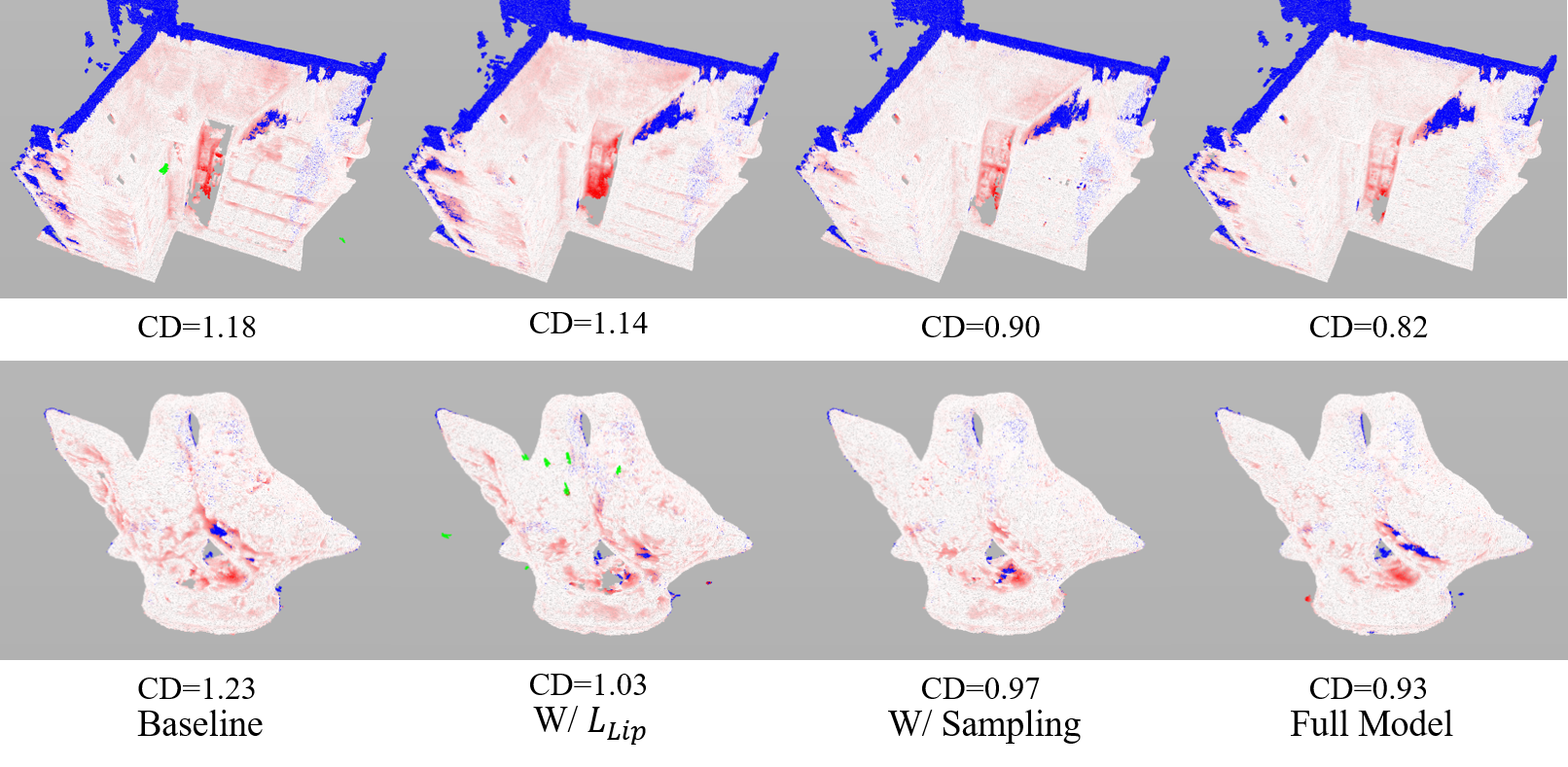}
\vspace{-0.4cm}
\caption{
Error map visualization.
White-to-red denotes increasing error, green marks distances beyond the cutoff, and blue indicates points outside the evaluated region.
}

  \label{fig:ablation}
\vspace{-0.4cm}
\end{figure}

\noindent\textbf{Impact of Lipschitz Regularization $L_{Lip}$. }
We first investigate the role of our Lipschitz constraint. As discussed in Section~\ref{sec:mlp_method}, imposing Lipschitz regularization $L_{Lip}$ in Eq.~\eqref{eq:LipConstraints} on the color network encourages smoothness in the implicit field. This smoothness ensures that high-frequency color variations in the input images correspond to significant geometric changes, rather than significantly being absorbed by the neural network's capacity, leading to richer geometric details, as shown by ``Baseline'' vs.~``w/$L_{Lip}$'' in Tab.~\ref{tab:ablation_horizontal}. Moreover, $L_{Lip}$ reduces floating artifacts caused by Gaussians overfitting appearance attributes at incorrect geometric positions, improving the CD from 1.29 to 1.23.

\noindent\textbf{Impact of Self-supervised Sampling. }
Our sampling strategy plays a crucial role in surface smoothing and refinement. By enforcing the color network to output the ground-truth color at the estimated surface depth (mean depth) while predicting zero color for points sampled before and behind the surface, we create a strong spatial constraint. To minimize the rendering error under this supervision, the MLP implicitly guides the Gaussians to concentrate around the mean depth and increases their opacity to form a solid surface. This concentration effect eliminates fuzzy boundaries and results in a much smoother and cleaner reconstruction, significantly lowering the CD to 1.17, as shown by ``w/Sampling'' in Tab.~\ref{tab:ablation_horizontal}. Moreover, we also justify that the supervision from both of the sampling before and after the estimated surface can improve the geometry inference.
The full model combines both strategies and achieves the best result, with CD reduced to 1.13 in Tab.~\ref{tab:ablation_horizontal}. Overall, the Lipschitz regularization improves detail and suppresses overfitting, while sampling supervision promotes surface smoothness.

\begin{table}[t]
\centering
\begin{minipage}{0.40\linewidth}
\centering
\caption{Impact of hash table capacity on geometry reconstruction on DTU~\cite{aanaes2016large}.}
\label{tab:hash_capacity}
\small
\resizebox{\linewidth}{!}{%
\begin{tabular}{c|cccc}
\toprule
Hash Size &$2^9$&$2^{12}$&$2^{15}$&$2^{18}$ \\
\midrule
CD $\downarrow$& 1.41 & 1.14 & \textbf{1.10} & 1.13 \\
\bottomrule
\end{tabular}
}
\end{minipage}
\hfill
\begin{minipage}{0.58\linewidth}
\centering
\caption{Ablation on sampling strategies.}
\label{tab:sampling_strategy}
\small
\resizebox{\linewidth}{!}{%
\begin{tabular}{c|cccc}
\toprule
Sampling Strategy &Surface&Surface+Front&Surface+Back&Full \\
\midrule
CD $\downarrow$&1.62&1.31&1.18&\textbf{1.13} \\
\bottomrule
\end{tabular}
}
\end{minipage}
\vspace{-0.8cm}
\end{table}
\noindent\textbf{Hash Table Capacity.}
We first investigate the effect of the hash table capacity on geometry reconstruction. 
As shown in Table~\ref{tab:hash_capacity}, increasing the hash size from $2^9$ to $2^{15}$ significantly improves the reconstruction accuracy, as a larger table provides higher feature capacity and reduces hash collisions. 
However, further increasing the size to $2^{18}$ slightly degrades the performance, which may be caused by over-parameterization and reduced regularization. A capacity of $2^{15}$ achieves the lowest CD in this ablation.

\noindent\textbf{Sampling Strategy.}
We further analyze different sampling strategies around the surface. 
As shown in Table~\ref{tab:sampling_strategy}, using only surface samples leads to inferior reconstruction quality. 
Introducing additional samples in front of the surface significantly improves the performance, as these samples provide constraints for the free space. 
Sampling behind the surface further improves the reconstruction accuracy by introducing interior occupancy information. 
The best performance is achieved when both front and back samples are used, demonstrating the effectiveness of modeling both free space and interior regions.

\noindent\textbf{View-Dependent vs. Non-View-Dependent. }We model the view-dependent color field with an MLP for high-fidelity rendering. Removing the ray direction leaves the Gaussian locations to be the only factor that determines the color, which hurts the rendering, highlighting its impact. As shown in Tab.~\ref{tab:view_dependence}, view-dependent color fields outperform non-view-dependent ones in both rendering and geometry inference.

\begin{wraptable}{r}{0.5\linewidth}
  \centering
  \vspace{-8pt}
  \caption{\textbf{Depth Update Interval.} Impact of update frequency on geometry.}
  \label{tab:ablation_interval}
  \small 
  \setlength{\tabcolsep}{4pt} %
  \begin{tabular}{c|cccc}
      \toprule
      Interval & 1 & 100  & 300 & 500 \\
      \midrule
      CD $\downarrow$ & \textbf{1.13} & 1.54 & 1.51 & 1.53 \\
      \bottomrule
  \end{tabular}
\vspace{-6pt}
\end{wraptable}
\vspace{-0.4em}
\noindent\textbf{Estimated Depth Update Interval.}
We study how often the estimated depth is updated for self-supervision. Tab.~\ref{tab:ablation_interval} compares the performance with update intervals of $\{1,100,300,500\}$ iterations. More frequent updates yield better performance, because the MLP observes richer coordinate variations despite noisy early estimates.

\begin{figure}[!h]
\vspace{-0.3cm}
  \centering

  \includegraphics[width=0.95\linewidth]{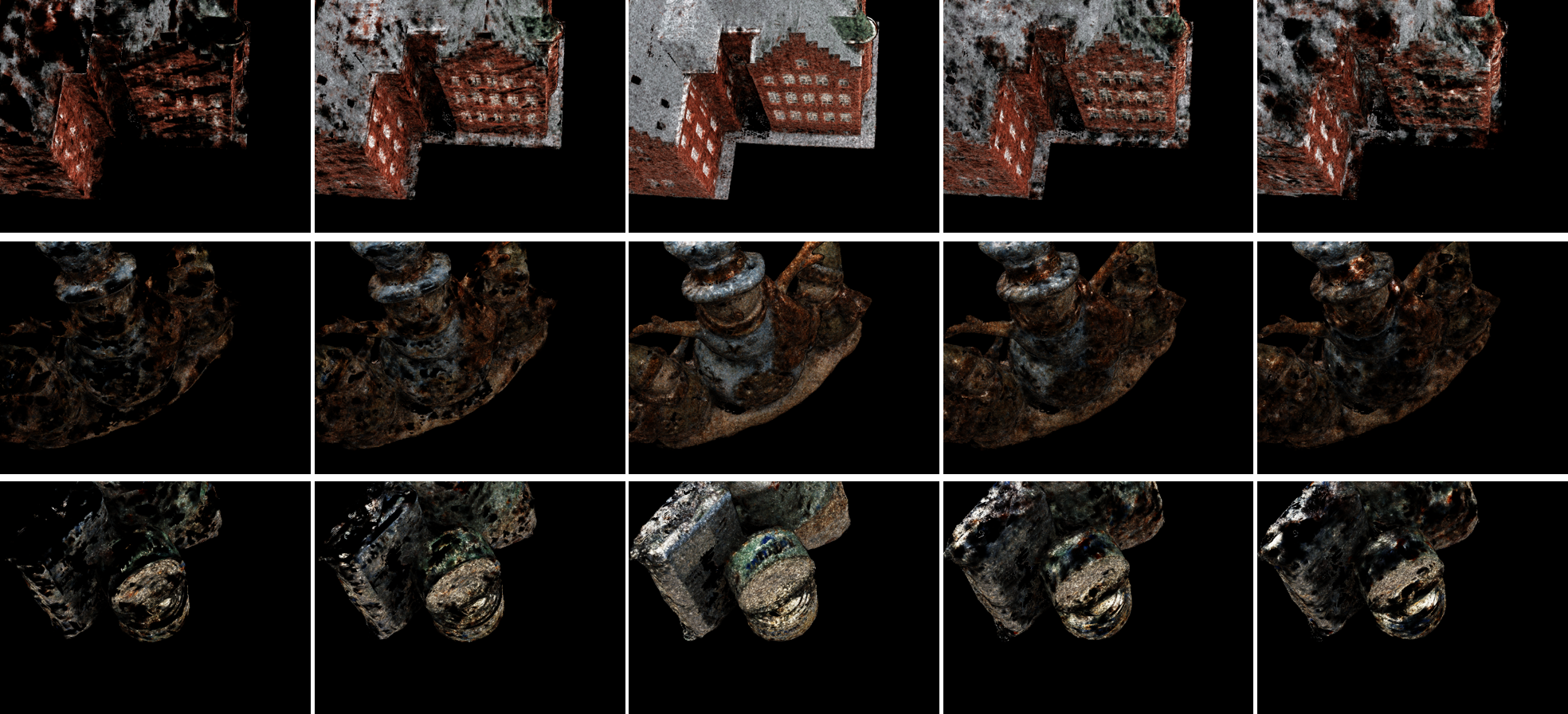}
  
  \caption{Visualization of the view-dependent color field. From left to right, we show MLP-predicted colors on a 2D slice moving from the front to the back of the object along the viewing direction.
  }
  
  \label{fig:vis_field}
\vspace{-0.3cm}
\end{figure}

\vspace{-0.4em}
\noindent\textbf{Visualization of View-Dependent Color Fields. }
Fig.~\ref{fig:vis_field} visualizes the learned view-dependent color field on a 2D slice under a fixed view direction. We sample 3D points around the rendered depth and query their colors with the trained MLP. The colors fade rapidly away from the surface, confirming the correctness of the learned field. 

\vspace{-0.4em}
\begin{wraptable}{r}{0.6\textwidth}
  \vspace{-0.2cm}
  \centering
  \caption{Color Fields vs. Opacity Fields.}
  \label{tab:ablation_color_vs_opacity}
  \begin{tabular}{l|ccc}
    \toprule

      Method & Ours(color) & Ours(opacity) & GOF~\cite{yu2024gof} \\
    \midrule
    CD $\downarrow$ & 0.77 & 2.94 & 0.74 \\

  \bottomrule
    \end{tabular}
\vspace{-0.2cm}
\end{wraptable}

\noindent\textbf{Color Fields vs.\ Opacity Fields. }
We also evaluate the modeling of opacity with an MLP. As reported in Tab.~\ref{tab:ablation_color_vs_opacity}, unlike color-field modeling, this alternative does not improve geometry. While GOF~\cite{yu2024gof} also models opacity and performs well, it relies on specialized rendering and additional normal and densification constraints, highlighting the simplicity and effectiveness of our color-field formulation.

\noindent\textbf{Difference to More Complex Constraints on Geometry. }
We show that continuous color fields are key to geometry recovery in 3DGS. Prior works add stronger geometric constraints (e.g., multi-view depth consistency, monocular depth/normal priors) at the cost of heavier computation. Instead, we provide a simpler and more general solution for geometry inference.

\vspace{-0.4em}
\noindent\textbf{More Ablation Studies. }
Further discussion of the Lipschitz-only comparison and ablations on front and back sampling variances are provided in Appendix~\ref{sec:ablation}.
\vspace{-0.2cm}

\section{Conclusion}
\vspace{-0.4cm}
We propose to constrain smoothness in radiance field for improving geometry inference with 3DGS. We model the radiance field within a neural view-dependent color field that is parameterized by a neural network. By leveraging spectral bias of neural network and Lipschitz regularization, our method encourages geometry to better align with color variation while maintaining the rendering quality. Our self-learning strategy requires no external priors and improves the learning of the color field. Extensive experiments on benchmarks demonstrate superior geometry reconstruction over recent 3DGS variants.
\vspace{-0.2cm}

\begin{ack}
\vspace{-0.2cm}
This work was partially supported by Deep Earth Probe and
Mineral Resources Exploration -- National Science and
Technology Major Project (2024ZD1003405) and the National
Natural Science Foundation of China (62272263).
\end{ack}
\par\vfill\par

\clearpage  %

\bibliographystyle{unsrtnat}
\bibliography{main}

\clearpage
\appendix
\renewcommand{\theHfigure}{A.\arabic{figure}}
\renewcommand{\theHtable}{A.\arabic{table}}

\renewcommand{\thefigure}{A.\arabic{figure}}
\renewcommand{\thetable}{A.\arabic{table}}
\setcounter{figure}{0}
\setcounter{table}{0}

\section{Overview}

This appendix provides additional implementation details, ablation studies, and discussions for our method. 
Our approach improves geometry inference in 3D Gaussian Splatting (3DGS~\cite{3dgs}) through a continuous neural view-dependent color field.

Section~\ref{sec:impl} presents additional implementation details of our framework. 
Section~\ref{sec:ablation} clarifies the Lipschitz-only comparison and analyzes the front and back sampling variances. 
Section~\ref{sec:discussion} discusses the limitations of our method and potential future research directions.

\section{Additional Implementation Details}
\label{sec:impl}
\subsection{The Color Field}

As mentioned in the framework, we utilize a 3D bounding box to delineate the foreground region. For 3D Gaussians situated within this bounding box, we employ our proposed neural color field (Hash Grid + MLP) to predict color attributes, whereas Gaussians outside this region continue to use standard Spherical Harmonics (SH) as the color representation. This design aims to ensure the hash grid and MLP can be trained with adequate 3D Gaussians.

This bounding box is not only used for color modeling but also limits the sampling for geometric supervision. During the optimization process, we restrict the sampling of surface points, as well as the asymmetric sampling of free-space points (before and behind the surface), exclusively to the interior of this bounding box.

\subsection{Details of Sampling Strategy}

Our method samples points on the estimated surface and in the free space for geometry inference. Here, we provide further implementation details regarding how these surface points are accurately derived from the volumetric rendering process.

\noindent\textbf{Depth Normalization.} 
In standard 3DGS~\cite{3dgs}, the expected depth for a pixel is rendered via volume rendering as $d = \sum_i w_i z_i$, where $w_i$ is the blending weight computed from the opacity $\alpha$ and transmittance. However, the accumulated opacity $\mathcal{A} = \sum_i w_i$ does not always sum to 1, particularly in regions representing the sky, object edges, or thin structures. This results in a rendered depth value that effectively includes a "zero-depth" contribution from the background, biasing the surface estimate closer to the camera than it actually is.

Inspired by 2DGS~\cite{2dgs}, we rectify this by normalizing the rendered depth before back-projection. The corrected depth $D'$ is computed as:
\begin{equation}
    D' = \frac{d}{\mathcal{A} + \epsilon},
\end{equation}
where $\epsilon$ is a small constant for numerical stability. This normalization ensures that the depth value corresponds to the actual geometric surface location, independent of the background's contribution.

\noindent\textbf{Validity Filtering.} 
To prevent supervising the network with unreliable geometric cues, we introduce a validity mask based on the accumulated opacity. We set a minimum opacity threshold $\tau = 0.1$. We only perform surface and free-space sampling on pixels where $\mathcal{A} > \tau$. This strategy filters out regions with low confidence, such as the sky or translucent artifacts at object boundaries, ensuring that our geometric supervision is applied only to solid, reliable surfaces.

\noindent\textbf{Back-Projection Logic.} 
Once valid pixels and their normalized depths $D'$ are obtained, we back-project them into 3D world space to generate the sampling points. Let $\mathbf{K}$ be the camera intrinsic matrix and $\mathbf{R}, \mathbf{t}$ be the camera rotation and translation (camera-to-world). For a pixel coordinate $(u, v)$, the ray direction in the world space, denoted as $\mathbf{r}_d$, is computed as,
\begin{equation}
    \mathbf{r}_d = \mathbf{R} \cdot \mathbf{K}^{-1} \cdot [u, v, 1]^T.
\end{equation}
The corresponding 3D surface point $\mathbf{x}_{surf}$ is then recovered by guiding the ray with the normalized depth,
\begin{equation}
    \mathbf{x}_{surf} = \mathbf{t} + D'(u,v) \cdot \mathbf{r}_d.
\end{equation}
These coordinates $\mathbf{x}_{surf}$ serve as the centers for our sampling strategy. We then sample points following a Gaussian distribution along the ray direction centered at $\mathbf{x}_{surf}$ to supervise the neural view-dependent color field.

\section{Additional Ablation Study}
\label{sec:ablation}
We provide further discussion of the contribution of Lipschitz regularization and additional ablations on the sampling variances used for front and back sampling on the DTU dataset~\cite{aanaes2016large}. The Chamfer Distance (CD) is used as the evaluation metric for geometry reconstruction.

\noindent\textbf{Contribution of Lipschitz Regularization}
\label{sec:lip_ablation}

Table~\ref{tab:ablation_horizontal} shows that Lipschitz regularization reduces averaged CD from 1.29 to 1.23 on DTU. Although sampling provides a larger gain, the two mechanisms
serve complementary roles: Lipschitz regularization encourages smoother color decoding, while sampling supplies spatial supervision. The reported full configuration achieves the lowest CD of 1.13, consistent with their intended complementary effects. The base and Lipschitz-only configurations differ only in whether Lipschitz regularization is enabled, all other settings are kept unchanged.

\begin{table}[h]
\centering

\begin{minipage}{0.48\linewidth}
\centering
\caption{Effect of front sampling variance.}
\label{tab:front_var}
\small
\begin{tabular}{c|ccc}
\toprule
Front Var. & 0.01&0.1&1.0\\
\midrule
 CD $\downarrow$&1.14 &\textbf{1.13}&1.57 \\
\bottomrule
\end{tabular}

\end{minipage}
\hfill
\begin{minipage}{0.48\linewidth}
\centering
\caption{Effect of back sampling variance.}
\label{tab:back_var}
\small
\begin{tabular}{c|ccc}
\toprule
Back Var. & 0.01&0.1&1.0\\
\midrule
CD $\downarrow$  &\textbf{1.13}&1.16&1.30  \\
\bottomrule
\end{tabular}

\end{minipage}

\end{table}
\noindent\textbf{Sampling Variance.}
We also study the influence of the sampling variance used for front and back sampling. 
As shown in Tables~\ref{tab:front_var} and \ref{tab:back_var}, a moderate variance achieves the best performance. 
When the variance is too small, the sampled points concentrate near the surface and provide limited spatial coverage. 
On the other hand, overly large variance introduces noisy samples that deviate far from the surface, which degrades the reconstruction accuracy. 
Based on these results, we set the front variance to $0.1$ and the back variance to $0.01$ in our final model.

\section{Discussion}
\label{sec:discussion}

Our method is designed as a complementary regularization for improving geometry inference in 3D Gaussian Splatting through a neural view-dependent color field. 
As such, its effectiveness depends on the amount of geometry supervision already present in the backbone, and its computational overhead depends on the relative training cost of the underlying Gaussian optimization.

\subsection{Analysis of Outdoor Scenes}
\label{sec:outdoor_analysis}

Table~\ref{tab:mipnerf360} shows a rendering-quality trade-off
on outdoor scenes.
For 3DGS, PSNR/SSIM/LPIPS change from
24.64/0.731/0.272 to 24.44/0.723/0.283,
with a similar trend for FastGS.
Our method primarily targets geometry refinement, and
Fig.~\ref{fig:mip_compare_visual} shows more complete
reconstructed surfaces in the displayed examples.
These observations suggest that improved surface completeness
can coexist with a small reduction in rendering quality.
However, without ground-truth geometry evaluation, the geometric
evidence remains qualitative, and rendering metrics alone do not
establish improved geometric accuracy.
The current comparison also does not isolate the cause of the
outdoor degradation, which remains a limitation of our method.

\subsection{Analysis of Highly Reflective Regions}
\label{sec:reflective_regions}

\begin{figure}[t]
\centering
\includegraphics[width=\linewidth]{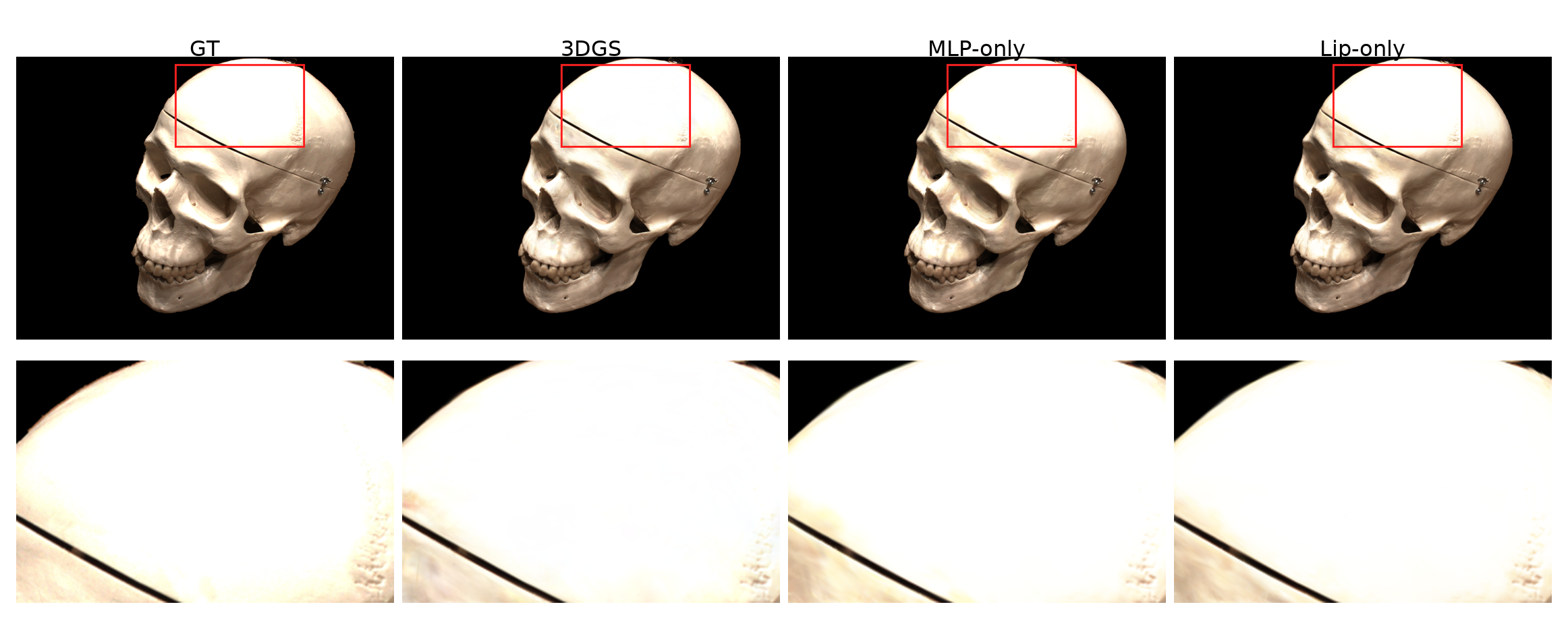}
\caption{Rendering comparison in a highly reflective region of DTU scan65. Columns show the ground-truth image, 3DGS, MLP-only, and Lip-only. The bottom row enlarges the regions marked by red boxes.}
\label{fig:scan65_reflective_render}
\end{figure}

\begin{figure}[t]
\centering
\includegraphics[width=\linewidth]{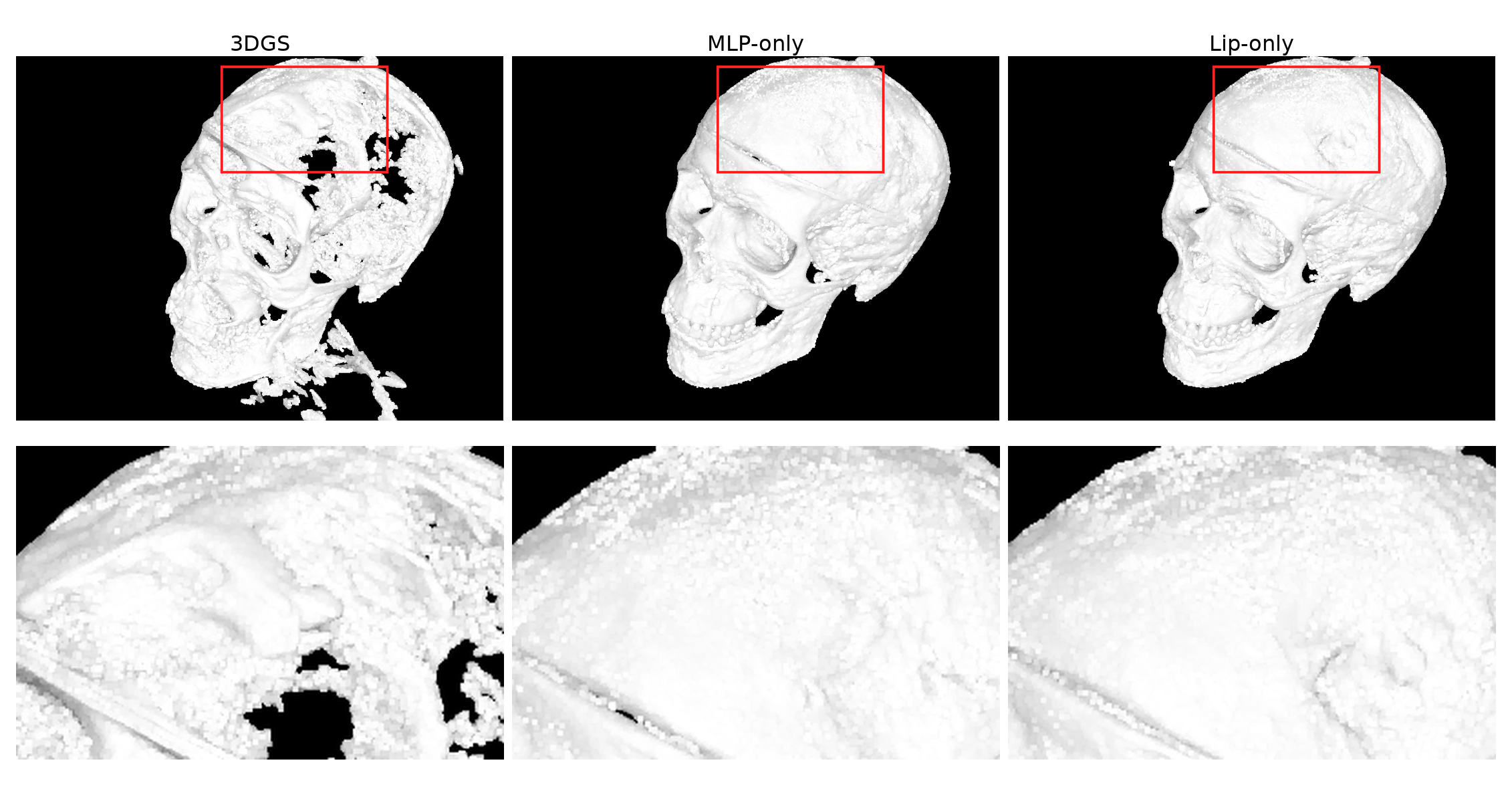}
\caption{Mesh comparison for DTU scan65. Columns show 3DGS, MLP-only, and Lip-only, with enlarged views of the highlighted region below. Both neural color field variants reduce the missing surface regions visible in the 3DGS reconstruction.}
\label{fig:scan65_reflective_mesh}
\end{figure}

We examine a highly reflective region in DTU scan65, comparing standard 3DGS, MLP-only, and Lip-only, where Lip-only adds Lipschitz regularization to MLP-only.
Figures~\ref{fig:scan65_reflective_render} and~\ref{fig:scan65_reflective_mesh} show the rendering and mesh comparisons, respectively.
Both neural color field variants recover more complete surfaces than 3DGS in the highlighted region.
Table~\ref{tab:scan65_reflective_roi} summarizes rendering quality in the selected image ROI and geometry quality in the corresponding 3D ROI.
Compared with MLP-only, Lip-only reduces local CD from 7.5207 to 7.2516 mm while improving PSNR, SSIM, and LPIPS.
Compared with 3DGS, Lip-only achieves lower CD and better SSIM and LPIPS, although its PSNR remains lower.
These results show that Lipschitz regularization can improve local geometry while retaining view-dependent appearance in this reflective example; they do not establish uniform improvements across all reflective regions.

\begin{table}[tbp]
\centering
\caption{Rendering and geometry quality in the selected highly reflective region of DTU scan65. Distance metrics are in millimeters; CD is the mean of the two directional distances. Best results are in bold.}
\label{tab:scan65_reflective_roi}
\small
\begin{tabular}{l|ccc|ccc}
\toprule
Method & PSNR$\uparrow$ & SSIM$\uparrow$ & LPIPS$\downarrow$ & s2d$\downarrow$ & d2s$\downarrow$ & CD$\downarrow$ \\
\midrule
3DGS & \textbf{30.4028} & 0.9538 & 0.1218 & 9.3344 & 18.1979 & 13.7661 \\
MLP-only & 29.1195 & 0.9567 & 0.1169 & 3.9021 & 11.1394 & 7.5207 \\
Lip-only & 29.3406 & \textbf{0.9589} & \textbf{0.1058} & \textbf{3.3971} & \textbf{11.1062} & \textbf{7.2516} \\
\bottomrule
\end{tabular}
\end{table}

\subsection{Compatibility with Strong Geometry-Constrained Backbones}

Our improvements are most visible on backbones with relatively weak or moderate geometry regularization, such as vanilla 3DGS~\cite{3dgs} and FastGS~\cite{fastgs}. 
For backbones that already impose strong geometry-oriented constraints, the room for further improvement is naturally smaller, because of the performance ceiling on the benchmark. 
This behavior is also consistent with our observations on methods such as 2DGS~\cite{2dgs} and PGSR~\cite{pgsr}. These approaches already incorporate strong geometric priors or multi-view consistent constraints, which partially overlap with the role of our sampling and Lipschitz-based regularization.
Therefore, when our method is integrated into such backbones, the additional gain may become marginal.

We emphasize that this does not indicate incompatibility. 
Rather, it reflects the intended role of our method as a lightweight complementary module rather than a replacement for heavily geometry-constrained pipelines. 
In other words, our method is most beneficial when the backbone still leaves optimization headroom for geometry refinement, while its marginal benefit decreases as the backbone itself becomes increasingly geometry-aware.

\subsection{Runtime Overhead on Fast Backbones}

Our method introduces a neural color field, implemented with a hash-grid encoder and an MLP, to predict view-dependent colors. 
This additional module incurs an extra training cost that is relatively stable across backbones. 
As a result, the relative runtime increase becomes more pronounced on highly accelerated methods such as FastGS~\cite{fastgs}. 
This is because FastGS substantially shortens the optimization time of the Gaussian backbone itself, making the training of the neural field account for a larger fraction of the total runtime.

Therefore, the larger slowdown observed on FastGS~\cite{fastgs} should be understood as a quality--efficiency trade-off rather than selective integration. 
Our method remains compatible with different 3DGS variants, but the practical trade-off differs depending on whether the backbone is bottlenecked by Gaussian optimization or by the added neural field. 
This also explains why the relative overhead is less severe on slower backbones such as vanilla 3DGS~\cite{3dgs}.

\subsection{Limitations}

Although our method improves the geometric fidelity of 3DGS, several limitations remain.

Our approach introduces an MLP to predict view-dependent colors instead of relying solely on the discrete color attributes of Gaussians. 
This design cannot be directly fused with the original rasterization pipeline and therefore introduces additional computational overhead during rendering. 
Moreover, the neural color field requires additional parameters, which increases the memory consumption compared to standard 3DGS~\cite{3dgs}.
\section{Code and Experimental Results}

We will organize and upload the code framework and experimental results
to our public repository:
\url{https://github.com/yangyixi21/Neural-Color-Fields-3DGS}.

\clearpage  %

\newpage
\section*{NeurIPS Paper Checklist}

\begin{enumerate}

\item {\bf Claims}
    \item[] Question: Do the main claims made in the abstract and introduction accurately reflect the paper's contributions and scope?
    \item[] Answer: \answerYes{} %
    \item[] Justification: The main claims in the abstract and introduction regarding the proposed method and its performance are accurately supported by the method described in Section~3 and the experimental results presented in Section~4.

\item {\bf Limitations}
    \item[] Question: Does the paper discuss the limitations of the work performed by the authors?
    \item[] Answer: \answerYes{} %
    \item[] Justification: The limitations of the work, including experimental constraints and assumptions, are discussed in the appendix.

\item {\bf Theory assumptions and proofs}
    \item[] Question: For each theoretical result, does the paper provide the full set of assumptions and a complete (and correct) proof?
    \item[] Answer: \answerNA{}
    \item[] Justification: The paper does not include theoretical results. The proposed method and loss function are introduced as a methodological design, validated experimentally in Section 4, without formal theorems or proofs.

    \item {\bf Experimental result reproducibility}
    \item[] Question: Does the paper fully disclose all the information needed to reproduce the main experimental results of the paper to the extent that it affects the main claims and/or conclusions of the paper (regardless of whether the code and data are provided or not)?
    \item[] Answer: \answerYes{}
    \item[] Justification: The paper fully discloses the information needed to reproduce the main results. The proposed method is described in Section~3, and the experimental details and hyperparameter settings are provided in Section~3 and the appendix.

\item {\bf Open access to data and code}
    \item[] Question: Does the paper provide open access to the data and code, with sufficient instructions to faithfully reproduce the main experimental results, as described in supplemental material?
    \item[] Answer: \answerNo{}
    \item[] Justification: The implementation is being prepared for public release at
\url{https://github.com/yangyixi21/Neural-Color-Fields-3DGS}.
The repository currently contains documentation only.
Method and experimental details are provided in Section~3
and the appendix.

\item {\bf Experimental setting/details}
    \item[] Question: Does the paper specify all the training and test details (e.g., data splits, hyperparameters, how they were chosen, type of optimizer) necessary to understand the results?
    \item[] Answer: \answerYes{}
    \item[] Justification: All key training and test details (e.g., data splits, hyperparameters, optimizer) are described in Section~3 and the appendix, with sufficient completeness to understand the results.

\item {\bf Experiment statistical significance}
    \item[] Question: Does the paper report error bars suitably and correctly defined or other appropriate information about the statistical significance of the experiments?
    \item[] Answer: \answerNo{}
    \item[] Justification: Due to the computational cost of 3D reconstruction experiments, results are reported from a single run with a fixed random seed, without error bars. The deterministic components of the pipeline are explicitly controlled, and the key comparisons are conducted under identical conditions.

\item {\bf Experiments compute resources}
    \item[] Question: For each experiment, does the paper provide sufficient information on the computer resources (type of compute workers, memory, time of execution) needed to reproduce the experiments?
    \item[] Answer: \answerYes{}
    \item[] Justification: The type of GPU, approximate runtime per experiment, and total compute used are reported in Section~4
    
\item {\bf Code of ethics}
    \item[] Question: Does the research conducted in the paper conform, in every respect, with the NeurIPS Code of Ethics \url{https://neurips.cc/public/EthicsGuidelines}?
    \item[] Answer: \answerYes{}
    \item[] Justification: The authors have reviewed the NeurIPS Code of Ethics and confirm that the research conforms to it in every respect. The paper uses only well-established public datasets and does not involve human subjects, sensitive data, or potentially harmful applications.

\item {\bf Broader impacts}
    \item[] Question: Does the paper discuss both potential positive societal impacts and negative societal impacts of the work performed?
    \item[] Answer: \answerNA{}
    \item[] Justification: This work is foundational research on a core 3D reconstruction algorithm and is not tied to specific downstream applications that would directly yield societal impacts.
    
\item {\bf Safeguards}
    \item[] Question: Does the paper describe safeguards that have been put in place for responsible release of data or models that have a high risk for misuse (e.g., pre-trained language models, image generators, or scraped datasets)?
    \item[] Answer: \answerNA{}
    \item[] Justification: The paper does not release models or data that pose a high risk for misuse. The proposed method is a 3D reconstruction technique that does not inherently involve generative models for disinformation or scraped datasets with safety concerns.

\item {\bf Licenses for existing assets}
    \item[] Question: Are the creators or original owners of assets (e.g., code, data, models), used in the paper, properly credited and are the license and terms of use explicitly mentioned and properly respected?
    \item[] Answer: \answerYes{}
\item[] Justification: All existing assets used in this work (DTU, Tanks and Temples, and Mip-NeRF360) are properly credited by citing the original papers that introduced them.

\item {\bf New assets}
    \item[] Question: Are new assets introduced in the paper well documented and is the documentation provided alongside the assets?
    \item[] Answer: \answerNA{}
    \item[] Justification: No implementation code, model checkpoints, or new datasets
are publicly released with this version.
The implementation and accompanying documentation
are being prepared for public release.

\item {\bf Crowdsourcing and research with human subjects}
    \item[] Question: For crowdsourcing experiments and research with human subjects, does the paper include the full text of instructions given to participants and screenshots, if applicable, as well as details about compensation (if any)? 
    \item[] Answer: \answerNA{}
    \item[] Justification: The paper does not involve crowdsourcing or research with human subjects. All experiments rely on publicly available benchmark datasets.

\item {\bf Institutional review board (IRB) approvals or equivalent for research with human subjects}
    \item[] Question: Does the paper describe potential risks incurred by study participants, whether such risks were disclosed to the subjects, and whether Institutional Review Board (IRB) approvals (or an equivalent approval/review based on the requirements of your country or institution) were obtained?
    \item[] Answer: \answerNA{}
    \item[] Justification: The paper does not involve crowdsourcing or research with human subjects, and therefore did not require IRB approval.

\item {\bf Declaration of LLM usage}
    \item[] Question: Does the paper describe the usage of LLMs if it is an important, original, or non-standard component of the core methods in this research? Note that if the LLM is used only for writing, editing, or formatting purposes and does \emph{not} impact the core methodology, scientific rigor, or originality of the research, declaration is not required.
    \item[] Answer: \answerNA{}
    \item[] Justification: The core method development in this research does not involve LLMs as any important, original, or non-standard components. LLMs were only used for minor writing polishing, which does not require declaration per NeurIPS policy.

\end{enumerate}
\end{document}